\documentclass[letterpaper, 10 pt, conference]{ieeeconf}
\IEEEoverridecommandlockouts
\usepackage{graphicx}
\usepackage{amsmath,amssymb,amsfonts}
\usepackage{algorithm,algpseudocode}
\newtheorem{property}{Property}

\newtheorem{condition}{Condition}
\newtheorem{proposition}{Proposition}

\usepackage{subcaption}
\usepackage{multirow}
\usepackage{booktabs}

\algdef{SE}[DOWHILE]{Do}{doWhile}{\algorithmicdo}[1]{\algorithmicwhile\ #1}%
\newcommand{\R}{\mathbb R}
\newcommand{\V}{\mathcal V}

\newcommand{\SDF}{\text{SDF}}
\DeclareMathOperator*{\argmin}{arg\,min}
\usepackage{multicol}
\begin{document}

\title{\LARGE \bf Narrow Passage Path Planning using  \\ Collision Constraint Interpolation}

\author{Minji Lee, Jeongmin Lee and Dongjun Lee$^\dagger$
\thanks{This research was supported by Samsung Research, the National Research Foundation (NRF) funded by the Ministry of Science and ICT (MSIT) of Korea (RS-2022-00144468), and the Ministry of Trade, Industry \& Energy (MOTIE) of Korea (RS-2024-00419641).}
\thanks{The authors are with the Department of Mechanical Engineering, IAMD and IOER, Seoul National University, Seoul, Republic of Korea, 08826.
\{mingg8,ljmlgh,djlee\}@snu.ac.kr. Corresponding author: Dongjun Lee. }
}

\maketitle
\thispagestyle{empty}
\pagestyle{empty}

\begin{abstract}
Narrow passage path planning is a prevalent problem from industrial to household sites, often facing difficulties in finding feasible paths or requiring excessive computational resources.
Given that deep penetration into the environment can cause optimization failure, we propose a framework to ensure feasibility throughout the process using a series of subproblems tailored for narrow passage problem.
We begin by decomposing the environment into convex objects and initializing collision constraints with a subset of these objects.
By continuously interpolating the collision constraints through the process of sequentially introducing remaining objects, our proposed framework generates subproblems that guide the optimization toward solving the narrow passage problem.
Several examples are presented to demonstrate how the proposed framework addresses narrow passage path planning problems.
\end{abstract}

\IEEEpeerreviewmaketitle 


\section{Introduction}

Path planning is a fundamental and crucial aspect of robotic tasks.
Path planning problem in narrow passage or cluttered environment remains particularly challenging with active researches having been conducted even to this day (\cite{ruan2022efficient, li2023sample, orthey2021section, hiraoka2024sampling}).
Such environments are not merely common in industrial settings, but also frequently encountered in household scenarios, such as tight assembly tasks or navigating cluttered spaces.

Historically, most path planning techniques have leaned on sampling-based methods such as rapidly-exploring random tree (RRT) \cite{lavalle1998rapidly} or probabilistic roadmap (PRM) \cite{kavraki1996probabilistic}.
While these methods offer certain advantages, such as convenient problem formulation and probabilistical completeness \cite{kavraki1998analysis}, they struggle in narrow passages, due to sampling inefficiencies \cite{li2023sample}. 

\begin{figure}
    \centering
    \includegraphics[width=8.8cm]{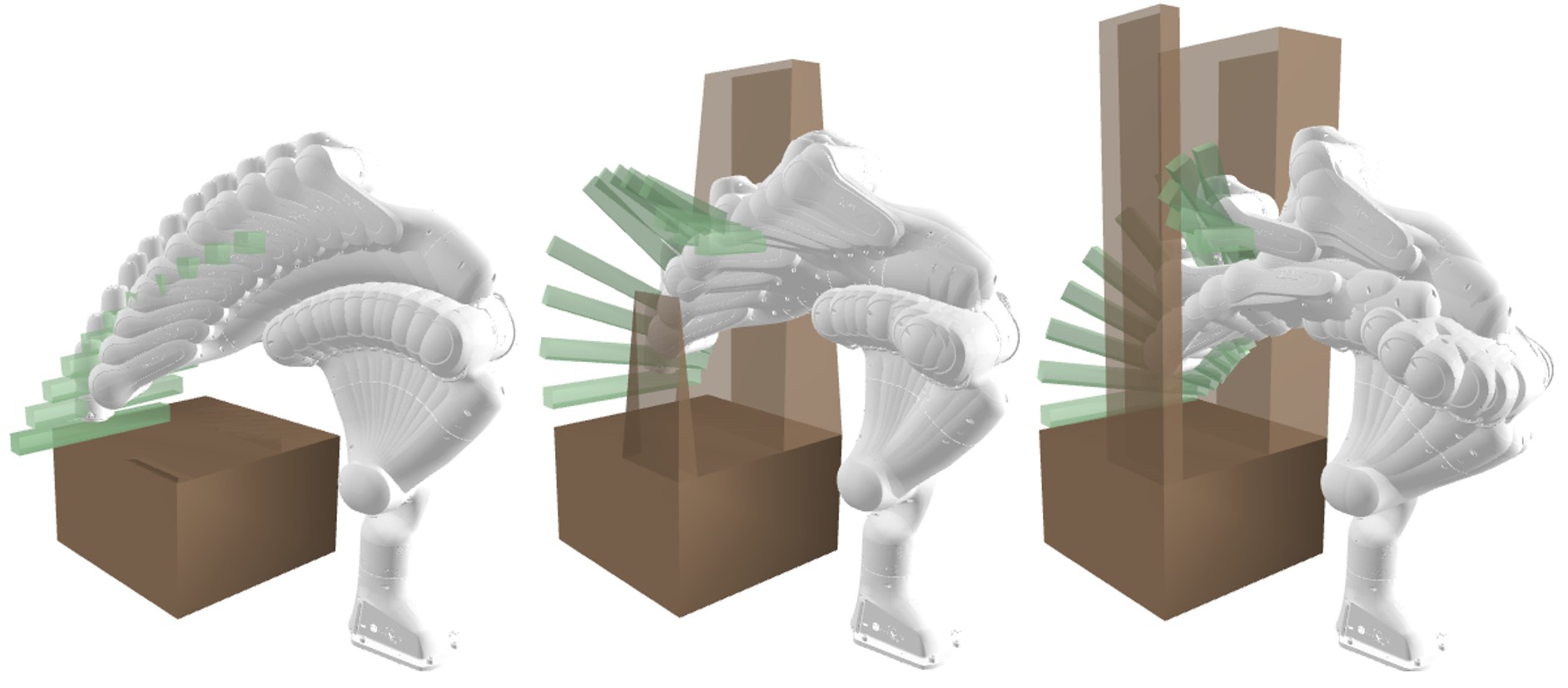}
    \caption{Optimization results for manipulator path planning during tool extraction from a narrow gap, utilizing the proposed collision constraint interpolation framework.}
    \label{fig:franka_tool}
\end{figure}

More recently, optimization-based path planning has emerged as another promising approach (\cite{ratliff2009chomp, dragan2011manipulation, schulman2014motion, kalakrishnan2011stomp}), formulating the path planning problem as an optimization.
These methods are able to quickly converge to paths with low cost in terms of path length, smoothness, or other task-specific metrics, by leveraging gradient information and, in some cases, second-order information. They also offer flexibility in integrating various cost and constraint factors.
Despite their advantages, as the problem is inherently non-convex when collision avoidance is considered, they can get stuck in local minima and fail to find an optimal or even feasible path without a sufficiently good initialization \cite{schulman2014motion}.

This difficulty is particularly pronounced in narrow passages, because deep penetration is more likely to occur.
There are two major factors that make the deep penetration problematic for trajectory optimization. First, contact features such as witness points, penetration depth, and contact normal are well-defined only when there is either no contact or minimal penetration \cite{erleben2018methodology}. 
Secondly, in situations of deep penetration, especially when the path passes through the medial axis \cite{pan2014predicting}, the contact normals between adjacent waypoints may become inconsistent, thus they may push the solutions in opposite directions, making it difficult to escape from infeasibility \cite{schulman2014motion}.
In essence, within these narrow passages, traditional path planning methods often fall short, struggling to find feasible paths or consuming excessive computational resources. 

In this paper, we propose addressing the problem through a novel optimization-based method with a series of subproblems tailored for narrow passage problems. 
We first relax the collision constraints so that we can initially start from an expanded free space, and, as the subproblems proceed, the constraints are gradually tightened back to their original form (See Fig. \ref{fig:franka_tool}).
This enables continuous refinement of the path, guiding it toward a feasible solution of the narrow passage problem while preventing deep penetration issues.

More specifically, we start by decomposing the environment into convex objects. 
Starting with a few initial objects, the remaining objects are then progressively introduced and augmented into the environment.
The addition is carried out through a certain interpolation between the Signed Distance Functions (SDFs) of newly-added and existing objects, in such a way that the path is continuously morphed into the final solution of the narrow passage problem without any tearing of the path, by considering the homotopy.

The paper is structured as follows: In Sec. \ref{sec:interpolation}, we introduce our collision constraint interpolation framework. Sec. \ref{sec:method1} details the optimization-based path planning algorithm using this collision constraint interpolation. Path planning examples and comparative analyses are presented in Sec. \ref{sec:resultandeval}, followed by conclusions in Sec. \ref{sec:conclusion}.

\section{Collision Constraint Interpolation} \label{sec:interpolation}

A key insight of our paper is that maintaining feasibility throughout optimization process is crucial for preventing the path from getting stuck in local minima, as it can avoid deep penetration, which is problematic for the optimization.
To achieve this, we propose a series of subproblems that initially simplify the collision constraints to create a wide free space, and then gradually revert it to the original collision constraint as the subproblems progress, while guiding the path to the solution of narrow passage problem.

\begin{figure}
    \centering
    \begin{subfigure}[b]{0.15\textwidth}
        \centering
        \includegraphics[width=2.5cm]{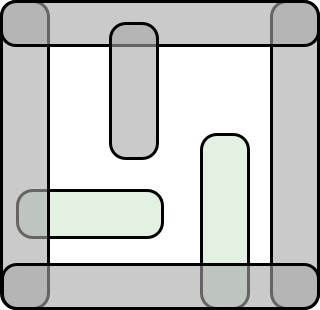}
        \caption{ }
        \label{fig:leafset1}
    \end{subfigure}
    \begin{subfigure}[b]{0.15\textwidth}    
        \centering
        \includegraphics[width=2.5cm]{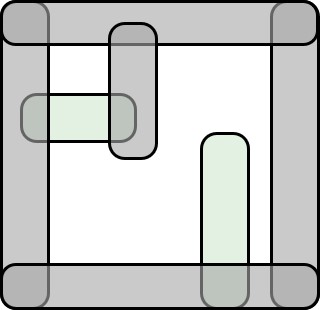}
        \caption{ }
        \label{fig:leafset2}
    \end{subfigure}
    \begin{subfigure}[b]{0.15\textwidth}    
        \centering
        \includegraphics[width=2.5cm]{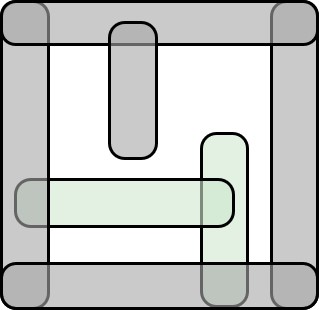}
        \caption{ }
        \label{fig:leafset3}
    \end{subfigure}
   \caption{Given an environment with convex objects $\V$ colored in gray, a set of green convex objects forms leaf set in (a), and not in (b) or (c). In (b), one of the green object violates Condition 1.1 by intersecting with two objects in $\V$. In (c), the green objects violate Condition 1.2 by intersecting with each other.} \label{fig:leafsets}
\end{figure}

To implement this framework, we first decompose the environment into a collection of convex objects $\V_{tot}=\{v_1, \cdots, v_{n_t}\}$, where $n_t$ is the number of the objects in $\V_{tot}$. Each object $v_i$ represents the space it occupies in the environment.
The environment is initialized with a subset of these objects $\V_{init} \subseteq \V_{tot}$.
Each subproblem is defined by gradually interpolating collision constraints of the sequence as new objects are progressively introduced into the environment.
At each sequence, a leaf set is introduced, satisfying following condition.
\begin{condition}[Leaf set] \label{condition} Let $\V^l$ be a set of convex objects with respect to a connected set of convex objects $\V$. The set $\V^l$ is a leaf set if it meets the following condition:
\begin{enumerate}
    \item Each element of $\V^l$ intersects with exactly one object from $\V$
    \item No elements in $\V^l$ intersect with each other.
\end{enumerate}
\end{condition}
Fig. \ref{fig:leafsets} illustrates a leaf set (Fig. \ref{fig:leafset1}) and counterexamples (Fig. \ref{fig:leafset2}, Fig. \ref{fig:leafset3}). Note that the gluing of the leaf set neither creates nor removes cycles, thereby preserving the homotopy of the environment.

To achieve interpolated collision constraints during the sequential addition of the leaf sets, we first examine the properties of SDF. Based on these properties, we define SDF interpolation between two convex objects, and propose an interpolation scheme to be applied across the sequences.


\subsection{Signed Distance Function}
A signed distance function (SDF) for an object $v$ is a function that quantifies the distance from any point $x \in \R^3$ to the surface of an object, formally defined as:
\begin{equation*}
    \SDF_v(x) = \begin{cases}
        -\inf\limits_{y \in \partial v} d(x-y) & \text{if } x \in v \\
        \inf\limits_{y \in \partial v} d(x-y) & \text{else}
    \end{cases}
\end{equation*}
where $\partial v$ is the boundary of $v$, and the metric $d$ is the commonly used Euclidean distance. 
For a collection of objects $\V=\{v_1, \cdots, v_n\}$, the combined SDF can be represented as:
\begin{equation*}
    \SDF_\V(x) := \min(\SDF_{v_1}(x), \cdots, \SDF_{v_n}(x))
\end{equation*}

Let us define an \textit{occupied space} of a function $g(\cdot):\R^3 \rightarrow \R$ as $\mathcal O(g) = \{x \in \R^3 ~|~g(x) \le 0 \}$.
If follows that the occupied space of a SDF of a set of objects, denoted as $\mathcal O_\V$, is exactly itself:
\begin{equation*}
    \mathcal O_\V := \mathcal O(\SDF_{\V}) = \bigcup_{v \in \V} v
\end{equation*}
Also, following is satisfied for any functions $g_1(\cdot), g_2(\cdot) \in \R^3\rightarrow \R$:
\begin{equation} \label{eq:occupied_min}
    \mathcal O(\min(g_1, g_2)) = \mathcal O(g_1) \cup \mathcal O(g_2)
\end{equation}


\subsection{Smoothed SDF Interpolation between Convex Objects}
\begin{figure}
        \centering
        \includegraphics[width=9cm]{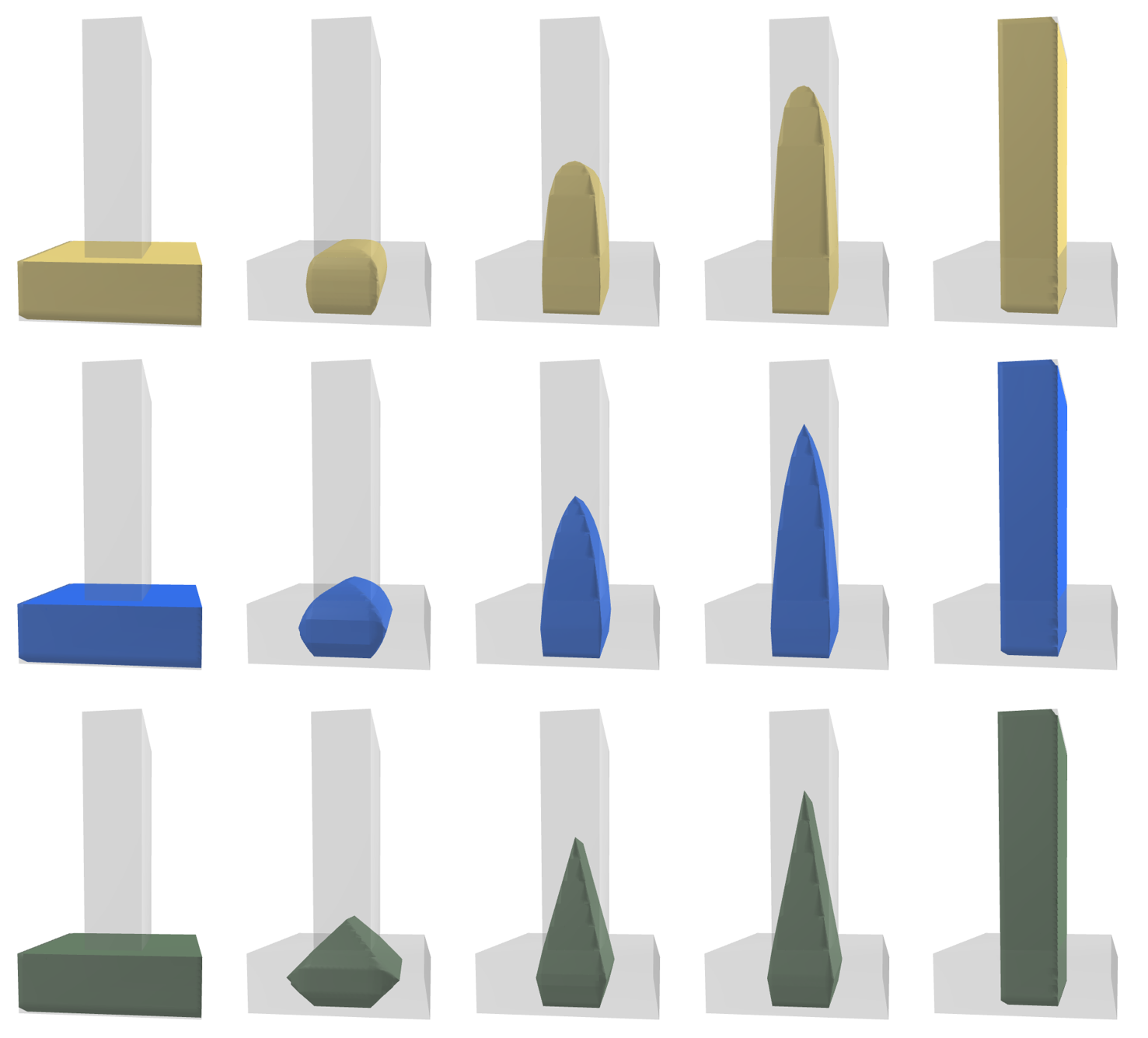}
    \caption{Object interpolation process using proposed shaping function \eqref{eq:exp_shaping} with $\eta = 40$ (top row), $\eta=15$ (middle row) and linear interpolation $\eta \rightarrow 0$ (bottom row).}
    \label{fig:interpolated_convex}
\end{figure}

We first define an \textit{interpolated SDF} between two intersecting convex objects $v_1$ and $v_2$ as:
\begin{equation} \label{eq:SDFinterp}
\SDF^\alpha_{v_1 \rightarrow v_2}(x) := (1-\alpha)f(\SDF_{v_1}(x)) + \alpha f(\SDF_{v_2}(x))
\end{equation}
where $\alpha \in [0, 1]$ is an interpolation variable and $f:\R\rightarrow \R$ is a shaping function that satisfies the following condition.
\begin{property}[Shaping function]
    A shaping function $f$ satisfies following conditions:
    \begin{enumerate}
        \item $f(0)=0$
        \item $f$ is non-decreasing, and convex
    \end{enumerate}  \label{prop:shaping}
\end{property}

Let us define an \textit{interpolated object}, $v^\alpha_{v_1 \rightarrow v_2}$, such that,
\begin{equation} \label{eq:interpolated_convex}
v^\alpha_{v_1 \rightarrow v_2} := \mathcal O\left(\SDF^\alpha_{v_1 \rightarrow v_2}\right)
\end{equation}
The following proposition outlines the properties of the interpolated object derived from any shaping function that satisfies Property \ref{prop:shaping}.
\begin{proposition} \label{th:convex}
    If two convex objects $v_1$ and $v_2$ intersect (i.e., $v_1 \cap v_2 \neq \varnothing$), their interpolated object $v^\alpha_{v_1\rightarrow v_2}$ as defined by \eqref{eq:interpolated_convex} is convex and satisfies:
    \begin{equation} \label{eq:convex_property}
        v_1 \cap v_2 \subseteq v^\alpha_{v_1 \rightarrow v_2} \subseteq v_1 \cup v_2
    \end{equation}
\end{proposition}
\vspace{3mm}
\begin{proof}
Since SDF of convex object is convex \cite{yan2020convexity}, both $\SDF_{v_1}(\cdot)$ and $\SDF_{v_2}(\cdot)$ are convex functions. Considering the properties of the shaping function, which are convex and non-decreasing, $f(\SDF_{v_1}(\cdot))$ and $f(\SDF_{v_2}(\cdot))$ also retain convexity.

Moreover, the function $\SDF^{\alpha}_{v_1 \rightarrow v_2}(\cdot)$, defined as a non-negative linear combination of the convex functions, inherently retains convexity.
Given that the level sets of a convex function are convex, it follows that the occupied space of $\SDF^\alpha_{v_1 \rightarrow v_2}$ is convex. Consequently, the interpolated object $v^\alpha_{v_1\rightarrow v_2}$ is also convex.

For any $x \in \R^3$ that is in $v_1\cap v_2$, following is satisfied:
\begin{align*}
     &~\SDF_{v_1}(x) \le 0~ \text{and} ~ \SDF_{v_2}(x) \le 0,\\
    \Rightarrow & ~\SDF^\alpha_{v_1 \rightarrow v_2} (x) \le 0
\end{align*}
Moreover for arbitrary $x \in v^\alpha_{v_1 \rightarrow v_2}$, following is satisfied:
\begin{align*}
     &~\SDF^\alpha_{v_1 \rightarrow v_2}(x) \le 0\\
    \Rightarrow & ~\SDF_{v_1} (x) \le 0~\text{or}~ \SDF_{v_2}(x) \le 0
\end{align*}
Therefore, $v_1\cap v_2 \subseteq v^\alpha_{v_1 \rightarrow v_2} \subseteq v_1\cup v_2$ is satisfied.
\end{proof}

The shaping function $f(\cdot)$ is designed to smooth the surface of the interpolated object.
For instance, if we do not use this shaping function (i.e. $f(x) = x$), some sharp ridges would be formed, which correspond to the medial axis \cite{turk2005shape} (See the bottom row of Fig. \ref{fig:interpolated_convex}).
These sharp ridges can cause abrupt changes in the contact normals between adjacent waypoints, which is highly undesirable for the optimization.

Such sharp ridges can be mitigated by using a properly-designed shape function $f(\cdot)$.
One example is modeled using an exponential formula, which adheres to Property \ref{prop:shaping}:
\begin{equation}
    f(x) = { 1 \over \eta }  \left( \exp \left( \eta x \right)-1 \right)\label{eq:exp_shaping}
\end{equation}
where $\eta \in \R^+$ acts as a scaling factor.
Fig. \ref{fig:interpolated_convex} visually demonstrates how varying values of $\eta$ affect the shape of the interpolated object. As $\eta$ increases, the interpolated object becomes increasingly smoothed, eliminating the sharp ridges.


\subsection{Homotopy Preserving Collision Constraint Interpolation}

\begin{figure}
        \centering
        \includegraphics[width=8.5cm]{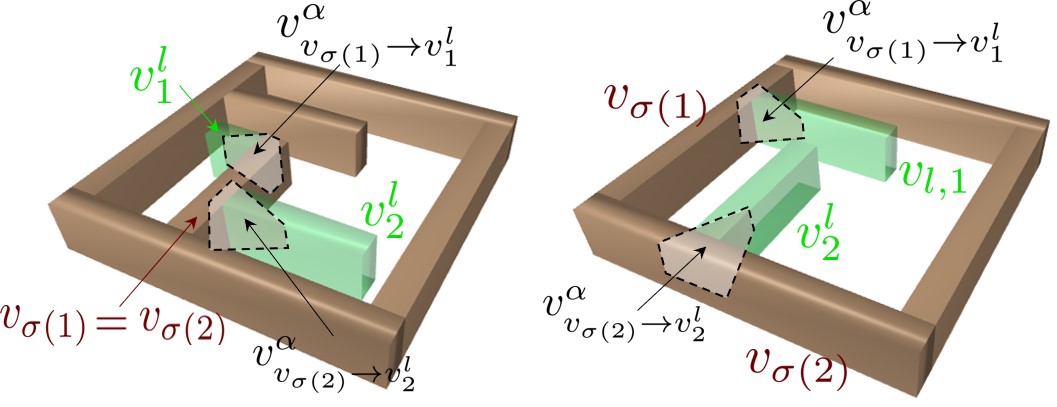}
    \caption{Illustration of two sequential processes of gluing leaf sets to the environment.}
    \label{fig:leafset_exp}
\end{figure}

Consider a $k$-th sequence of adding a leaf set $\V^l_k= \{v^l_1, \cdots, v^l_{n_l}\}$, where $n_l$ is the number of objects in the leaf set, into a current environment characterized by objects $\V_k=\{v_1, \cdots, v_n \} \subseteq \V_{tot}$, where $n$ is the number of objects in current environment. Note that $n$ and $n_l$ vary with the sequence $k$, but for brevity, we omit this dependence in the notation. Each convex object $v^l_j$ intersects uniquely with an object $ v_{\sigma(j)} \in \V_k$, where $\sigma(\cdot)$ maps the index of the intersecting object in $\V_k$. See Fig. \ref{fig:leafset_exp} for the graphical illustrations.
Then, this sequence can be interpolated using the interpolation variable $\alpha$ as:
\begin{equation}
    \SDF^\alpha_{\V_k+\V^l_k}(x) := \min \Big(\SDF_{\V_k}(x), \SDF^\alpha_1(x), \cdots, \SDF^\alpha_{n_l}(x)\Big) \label{eq:interpolated_free_space} 
\end{equation}
where $\SDF_j^\alpha(x) := \SDF^\alpha_{v_{\sigma(j)} \rightarrow v^l_{j}}(x)$.
When $\alpha=0$, the occupied space is exactly $\mathcal O_{\V_k}$, and at $\alpha=1$, it expands to $\mathcal O_{\V_k \cup \V^l_k}$:
\begin{equation*}
    \SDF^\alpha_{\V_k+\V^l_k} = \begin{cases}
        \SDF_{\V_k} & \text{if } \alpha=0\\
        \SDF_{\V_k \cup \V^l_k} & \text{if } \alpha=1
    \end{cases}
\end{equation*}

\begin{figure}
    \centering
    \begin{subfigure}[b]{0.47\textwidth}
        \centering
        \includegraphics[width=7.8cm]{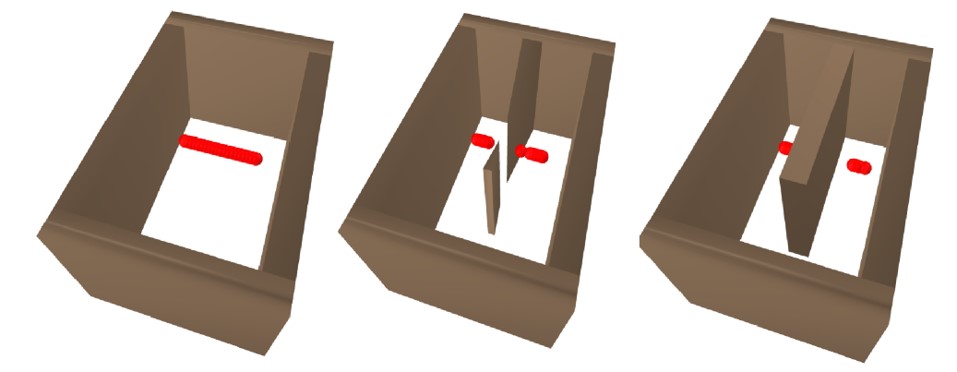}
        \caption{The path cannot stay within the free space with continuous updates, due to the change in homotopy.}
        \label{fig:sdf_interp_tearing}
    \end{subfigure}
    \\   \vspace{3mm}    
    \begin{subfigure}[b]{0.46\textwidth}    
        \centering
        \includegraphics[width=7.8cm]{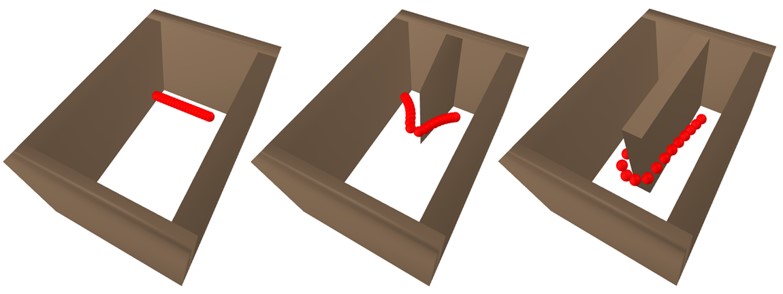}
        \caption{Using our framework, the path can be maintained in free space with continuous update, leading to successful path planning.}
        \label{fig:sdf_interp}
    \end{subfigure}
    \vspace{3mm}
   \caption{Comparison of homotopy equivalence and the corresponding path planning outcomes when performing interpolation using two distinct formulas.} \label{fig:particle_traj}
\end{figure}

As shown in Fig. \ref{fig:particle_traj}, preserving homotopy of the occupied space is essential during this interpolation; if lost, the path may stuck in an infeasibility, unable to retain within the free space through any continuous deformation (Fig. \ref{fig:sdf_interp_tearing}). Conversely, if homotopy is preserved, the path is more likely to be continuously updated to stay within the free space (Fig. \ref{fig:sdf_interp}). 

From \eqref{eq:occupied_min}, the occupied space of the interpolated collision constraint \eqref{eq:interpolated_free_space} can be characterized using the interpolated objects as:
\begin{align}
    \mathcal O \left( \SDF^\alpha_{\V_k+\V^l_k} \right) &=  \mathcal O_{\V_k} \cup \bigcup_{j=1}^{n_l} v^\alpha_{v_{\sigma(j)} \rightarrow v^l_j} \nonumber \\
    & =\mathcal O_{\V_k} \cup \bigcup_{j=1}^{n_l} \left\{ v^\alpha_{v_{\sigma(j)}\rightarrow v^l_j} \cap (v_{\sigma(j)} \cup v^l_j) \right\} \label{eq:occupied1} \\
    & =\mathcal O_{\V_k} \cup \bigcup_{j=1}^{n_l} \left\{ v^\alpha_{v_{\sigma(j)} \rightarrow v^l_j} \cap v^l_j \right\} \label{eq:occupied2}
\end{align}
The transformation to \eqref{eq:occupied1} is derived from the property \eqref{eq:convex_property}, while \eqref{eq:occupied2} is derived from the fact that $v_{\sigma(j)} \in \V_k$. Let us denote $v^\alpha_j := v^l_j \cap v^\alpha_{v_{\sigma(j)} \rightarrow v^l_j}$ for simplicity. Then, the occupied space \eqref{eq:occupied2} can be interpreted as a gluing of a set $\{v^\alpha_1, \cdots, v^\alpha_{n_l} \}$. The set $\{v^\alpha_1, \cdots, v^\alpha_{n_l}\}$ is a leaf set, since each $v^\alpha_j$ does not intersect with one another, and intersects with only $v_{\sigma(j)}$ among $\V_k$:
\begin{align*}
    &v^\alpha_i \cap v^\alpha_j \subseteq v^l_i \cap v^l_j = \varnothing\\
    &v^\alpha_j \cap v_{\sigma(j)} \supseteq v^l_j \cap v_{\sigma(j)} \neq \varnothing \\
    &v^\alpha_j \cap v_{\sigma(i)} \subseteq v^l_j \cap v_{\sigma(j)} = \varnothing
\end{align*}
for all $i\neq j$.
Since gluing of a leaf set do not create or delete cycles, thus the occupied space \eqref{eq:occupied2} preserves homotopy with $\V_k$ for all $\alpha \in [0, 1]$.
Consequently, the homotopy of the occupied space for the interpolated collision constraint \eqref{eq:interpolated_free_space} is preserved throughout the interpolation.

\subsection{Generation of Leaf Set Sequences}

\begin{algorithm}[t]
    \caption {Sequences of leaf sets}
    \begin{algorithmic}[1]
        \State \textbf{Input}: Decomposed objects $\V_{tot}$
        \State \textbf{Output}: Sequence of leaf sets $\V^l$, Initial objects $\V_{init}$
        \State $\V \gets \V_{tot}$
        \State $k \gets 1$
        \Repeat
            \State Initialize $\V^l_k = [~]$
            \For{$v \in \V$}
                \If{$\V^l_k \cup v$ satisfies Condition \ref{condition}}
                    \State Append $v$ to $\V^l_k$
                \EndIf
            \EndFor
            \State $\V \gets \V \setminus \V^l_k$
            \State $k \gets k+1$
        \Until{$|\V^l_k|=0$ or $|\V|=0$}
        \State $\V^l = \{\V^l_k, \V^l_{k-1}, \cdots, \V^l_1\}$
        \State $\V_{init} = \V$
    \end{algorithmic}
    \label{alg:sequence}
\end{algorithm}

The initial objects and the sequence of the additions can be determined automatically using Algorithm \ref{alg:sequence}. Starting with the total set of objects, the algorithm identifies a leaf set $\V^l_k$ at each iteration, where $k$ represents the current step of sequence. The leaf set is formed by adding objects which do not intersect with each other but uniquely intersect with $\V$ (Line 7-11). This subset $\V^l_k$ is then removed from $\V$ (Line 12), and the process repeats with the next value of $k$ (Line 13) until no further leaf sets can be identified. 
The final sequence of leaf sets is established by reversing the order of $\V^l_k$ (Line 15) about $k$. The remaining objects $\V$, after all leaf sets have been removed, become the initial objects $\V_{init}$ (Line 16).

This algorithm is designed to leave as few objects as possible as initial objects. However, if many objects need to be introduced sequentially, the optimization process may take longer. In such cases, the user can choose not to include all possible objects in the leaf set in Lines 7-11, which will not affect the subsequent path planning framework.
    
\section{Path Planning using \\ Collision Constraint Interpolation} \label{sec:method1}

\begin{algorithm}[t]
    \caption {Narrow passage path planning}
    \begin{algorithmic}[1]
        \State \textbf{Input}: Total set of nodes $\V_{tot}$
        \State Determine $\V^l, \V_{init}$ from Alg. \ref{alg:sequence}
        \State Initialize environment $\V_1 = \V_{init}$ and path $X_{1:T}$
        \For{$k=1,\cdots, |\V^l|$}
            \State $\alpha = 0$
            \While{$\alpha < 1$}
                \State Refine the path $X_{1:T}$ by optimization \eqref{eq:subproblem}
                \State $\alpha \gets \min(\alpha + \Delta \alpha, 1)$
            \EndWhile
            \State $\V_{k+1} = \V_k \cup \V^l_k$
        \EndFor
        \State \textbf{return} $\V_{seq}$
        \end{algorithmic}
    \label{alg:overall}
\end{algorithm}

A typical path planning optimization can be formulated as:
\begin{equation} \label{eq:optimization_final}
\begin{gathered}
 \min_{X_{1:T}} \sum_{t=1}^{T-1} \|X_{t+1} - X_t \|^2 \\
\text{s.t. } \min_{x \in W(X_t)} \SDF_{\V}(x) \ge \hat d, ~t=1, \cdots,T
\end{gathered}
\end{equation}
where $X_{1:T}$ is the optimization variable representing configuration at the $t$-th timestep, $T$ is the total number of waypoints, $W(X_t)$ is the surface of the robot at each waypoint $X_t$, and $\hat d \in \R^+$ is the predefined safe distance.
Here, we aim to generate a series of subproblems tailored to narrow passage problem using the collision constraint interpolation in Sec. \ref{sec:interpolation}.

The overall path planning algorithm is explained in Algorithm \ref{alg:overall}.
We first decompose the environment into a set of convex objects, and initialize the path by ignoring the constraint of \eqref{eq:optimization_final}. Then, the leaf set addition sequence and initial objects can be identified using Algorithm \ref{alg:sequence}.
Then, the subproblems are defined by substituting the collision avoidance constraint in \eqref{eq:optimization_final} with the interpolated collision constraint:
\begin{equation} \label{eq:subproblem}
\begin{gathered}
 \min_{X_{1:T}} \sum_{t=1}^{T-1} \|X_{t+1} - X_t \|^2 \\
\text{s.t. } \min_{x \in W(X_t)} \SDF^\alpha_{\V_k+\V^l_k}(x) \ge \hat d
\end{gathered}
\end{equation}
For each $k$-th sequence, $\alpha$ is gradually increased from 0 in increments of $\Delta \alpha$, creating a series of subproblems.
In each subproblem, the collision constraint becomes progressively stricter, guiding the optimization toward a solution to the narrow passage planning problem \eqref{eq:optimization_final}.

To solve the subproblem \eqref{eq:subproblem}, we employ Sequential Quadratic Programming (SQP) by linearizing the problem. The resulting Quadratic Programming (QP) at each iteration is solved using SubADMM \cite{lee2023modular}, which is particularly adept at stably and efficiently solving conflicting constraints common in narrow passage path planning.

We need to detect the collision to solve the problem \eqref{eq:subproblem}. 
Even though the environment changes with $\alpha$, by leveraging the SDF interpolation, we can efficiently manage the collision detection.
By separating the interpolated collision constraint \eqref{eq:interpolated_free_space} into two parts, we can express the constraint as follows:
\begin{align}
    &\text{SDF}_{\V} (x) \ge \hat d \label{eq:collision_prev}\\
    &\text{SDF}^\alpha_{v_{\sigma(j)} \rightarrow v^l_j}(x) \label{eq:collision_interp}\ge \hat d, ~j \in \{1, \cdots, m\}
\end{align}
where \eqref{eq:collision_prev} represents the collision constraint for the current environment $\V$ and \eqref{eq:collision_interp} corresponds to the collision constraint for the interpolated convex objects.

The collision detection for the interpolated convex objects \eqref{eq:collision_interp} can be performed using the Frank-Wolfe algorithm \cite{macklin2020local}.
That is, from the given triangular mesh of the robot $W(X_t)=\{W_1, \cdots, W_{n_w}\}$ at waypoint $X_t$, collision detection can be conducted as:
\begin{equation*}
    x^* = \argmin_{x \in W_i} \text{SDF}^\alpha_{v_{\sigma(j)} \rightarrow v^l_j}(x)
\end{equation*}
where $W_i$ is the $i$-th triangular mesh. 
Due to the convexity of the interpolated SDF \eqref{eq:SDFinterp}, the solution converges to the global optimum at a sublinear rate.
One key advantage of employing SDF-based interpolation lies in its ability to handle collision detection without requiring the computation of the geometry of the interpolated environment.

Unlike the interpolated convex objects, where the geometry changes with $\alpha$, the objects in $\V$ have fixed geometries that can be represented with meshes. This allows for faster collision detection methods beyond Frank-Wolfe such as the Gilbert-Johnson-Keerthi with Expanding Polytope Algorithm (GJK-EPA) \cite{van2001proximity}.

\section{Evaluations} \label{sec:resultandeval}
We test our framework in three different scenes, 1) placing dishes on a drying rack, 2) extracting a box from a narrow gap, and 3) maze navigation of nonhonolonomic system.
The first two scenarios are compared against the results obtained using baseline planners from TrajOpt \cite{schulman2014motion}, OMPL \cite{coleman2014reducing} and CHOMP \cite{ratliff2009chomp}.
The timeout for the OMPL was set to 30 seconds and 50 seconds for each respective scenario.
\subsection{Placing Dishes on the Rack}
\begin{figure}
    \centering
    \includegraphics[width=8.9cm]{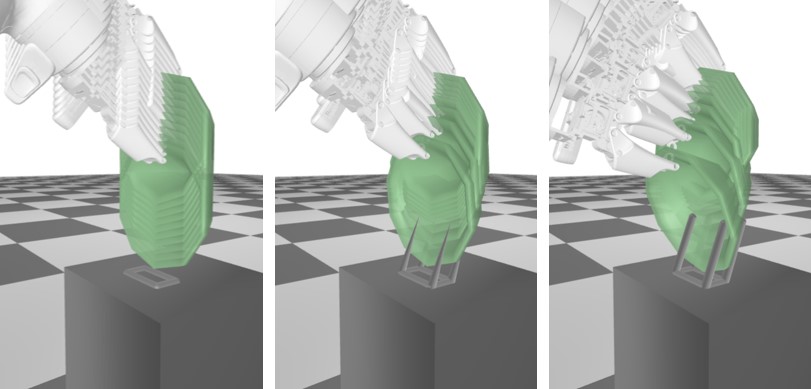}
    \caption{Results of a successful path planning using the proposed framework for the dish insertion task.}
    \label{fig:enter-label}
\end{figure}

Inserting a dish into a narrow gap of a drying rack is challenging for a manipulator \cite{lee2023uncertain}. The thinness of the rack makes deep penetration more likely, resulting in being stuck in infeasibility. For the same reason, even achieving the feasible goal position is challenging for this scenario. 

We define the objective function as the distance to an approximate reference pose located at the center of the rack, facing horizontal direction.
Additionally, the Cartesian path length objective function and a hard constraint on the initial joint position are incorporated.
By employing our proposed initialization scheme and refinement process, feasible final position of the plate placed on the drying rack, along with a feasible path could be achieved.

Table \ref{tb:dish_ablation} compares the result of the optimization with and without collision constraint interpolation. Success time shows the mean and standard deviation of the elapsed time of success cases, while total time shows both success and failure cases.
The tests are conducted using combinations of three different shapes of dishes and 20 different racks. The results indicate that our method outperforms the one without collision constraint interpolation in both success rate and computation time.

\begin{table}[h]
\centering
    \begin{tabular}{ c c c c }
    \toprule
     Method & \textbf{Success} & \textbf{Total time (s)} & \textbf{Success time (s)} \\ 
     \midrule \midrule
     Proposed & 58/60 & 3.56 $\pm$ 0.69 & 3.95 $\pm$ 0.68 \\  
     Without interpolation & 23/60 & 7.06 $\pm$ 0.33 & 7.42 $\pm$  0.27 \\
     \bottomrule
    \end{tabular}
    \caption{Ablation study of dish placing with and without collision constraint interpolation.}
    \label{tb:dish_ablation}
\end{table}

\begin{table}[ht]
    \centering   
     \begin{tabular}{@{}cccc@{}}
        \toprule
        \textbf{Method} & \textbf{Success} & \textbf{Total time (s)} & \textbf{Success time (s)} \\ 
         \midrule \midrule
        Proposed & 20/20 & 3.28 $\pm$ 0.61 &  3.28 $\pm$ 0.61  \\
        \midrule
        RRTConnect & 9/20  &  20.08 $\pm$ 11.18 & 10.45 $\pm$ 10.32  \\
        BiTRRT & 10/20 & 22.21 $\pm$ 9.91 & 16.24 $\pm$ 11.10  \\
        TRRT & 2/20 & 26.37 $\pm$ 9.02  & 1.36 $\pm$ 0.00 \\
        BiEST & 4/20 & 23.42 $\pm$ 9.91 & 16.24 $\pm$ 11.10 \\
        BMFT & 2/20 & 21.34 $\pm$ 7.86 & 10.85 $\pm$ 8.33 \\
        PRMstar & 0/20 & 31.67 $\pm$ 0.90 & - \\
        LazyPRM & 0/20 & 30.03 $\pm$ 0.00 & - \\
        KPIECE & 1/20 & 22.20 $\pm$ 9.09 & 8.48 $\pm$ 0 \\
        BKPIECE & 4/20 & 21.34 $\pm$ 7.86 & 10.85 $\pm$ 8.33 \\
        \midrule
        TrajOpt & 0/20 & 4.76 $\pm$ 1.27 &  - \\
        CHOMP & 2/20 & 22.74 $\pm$ 2.58 &  15.28 $\pm$ 0.03\\
        \bottomrule
    \end{tabular}
    \caption{Comparison with other methods for the task of placing dish, showing the planning time and success rate, with the goal position provided by our framework.}
    \label{tab:dish_baselines}
\end{table}

Conventional planning methods typically require a predefined goal pose. To compare our proposed framework with existing methods, we used the feasible final pose obtained by our framework as the goal position for the other methods. We tested twelve different shapes of dish racks, with the results presented in Table \ref{tab:dish_baselines}.
As also shown in \cite{li2023sample}, conventional sampling-based methods—except for BiTRRT \cite{devaurs2013enhancing}—faced significant challenges in solving the narrow passage problem.
Despite making the problem easier for the baselines by providing the goal pose, our method still outperformed all others in both success rate and computation time. 
Note that in the case of TrajOpt, continuous collision detection is performed by assuming a convex hull between waypoints. However, this results in an overly conservative over-approximation for the non-convex geometry of the dish, leading to failure in successful execution.

\subsection{Extraction of Tool from a Narrow Gap}

\begin{table}[t]
    \centering   
     \begin{tabular}{@{}cccc@{}}
        \toprule
         \textbf{Method} & \textbf{Success} & \textbf{Total time (s)} & \textbf{Success time (s)}\\ 
         \midrule \midrule
        Proposed &  30/30 & 6.52 $\pm$ 0.69  & 6.52 $\pm$ 0.69 \\
        \midrule
        RRTConnect & 1/30 & 48.83 $\pm$ 3.54 & 32.91 $\pm$ 0.00  \\
        BiTRRT & 19/30 & 30.33 $\pm$ 14.23 & 23.15 $\pm$ 12.80 \\
        TRRT & 0/30 & 50.03 $\pm$ 0.00 & - \\
        BiEST & 0/30 & 50.07 $\pm$ 0.16 & -  \\
        BMFT & 1/30 & 49.51 $\pm$ 4.98 & 32.41 $\pm$ 0.00  \\
        PRMstar & 0/30 & 50.28 $\pm$ 0.41 & - \\
        LazyPRM &  0/30 & 50.04 $\pm$ 0.01 & - \\
        KPIECE & 0/30 & 50.04 $\pm$ 0.015 &  - \\
        BKPIECE & 0/30 & 50.07 $\pm$ 0.14 & - \\
        \midrule
        TrajOpt & 17/30  & 3.41 $\pm$ 1.09 & 2.71 $\pm$ 0.93 \\
        CHOMP & 0/30 & 32.66 $\pm$ 11.20 & -  \\
        \bottomrule
    \end{tabular}
    \caption{Comparison with other methods for the task of extracting tool from a narrow gap, showing the planning time and success rate.}
    \label{tab:table1}
\end{table}
Taking tool out through a narrow gap is a challenge task for a manipulator, especially when the size of the tool is large, or the obstacles are thin.
Our objective is to optimize the manipulator path, starting from a pose that grasping the object, and extracting it out through a narrow gap. Fig \ref{fig:franka_tool} shows the result of the optimization.

We introduced slight randomness into the environment configuration to create 30 environments. Each planner was tested under these conditions.
As shown in Table \ref{tab:table1}, CHOMP and sampling methods had low success rates and longer planning times compared to our approach. While TrajOpt achieved higher success rates and shorter planning times than the sampling methods, its success rates were still lower than our framework, which successfully planned in all configurations.

\subsection{Nonholonomic Navigation of Maze}

\begin{figure}
    \centering
    \includegraphics[width=8.4cm]{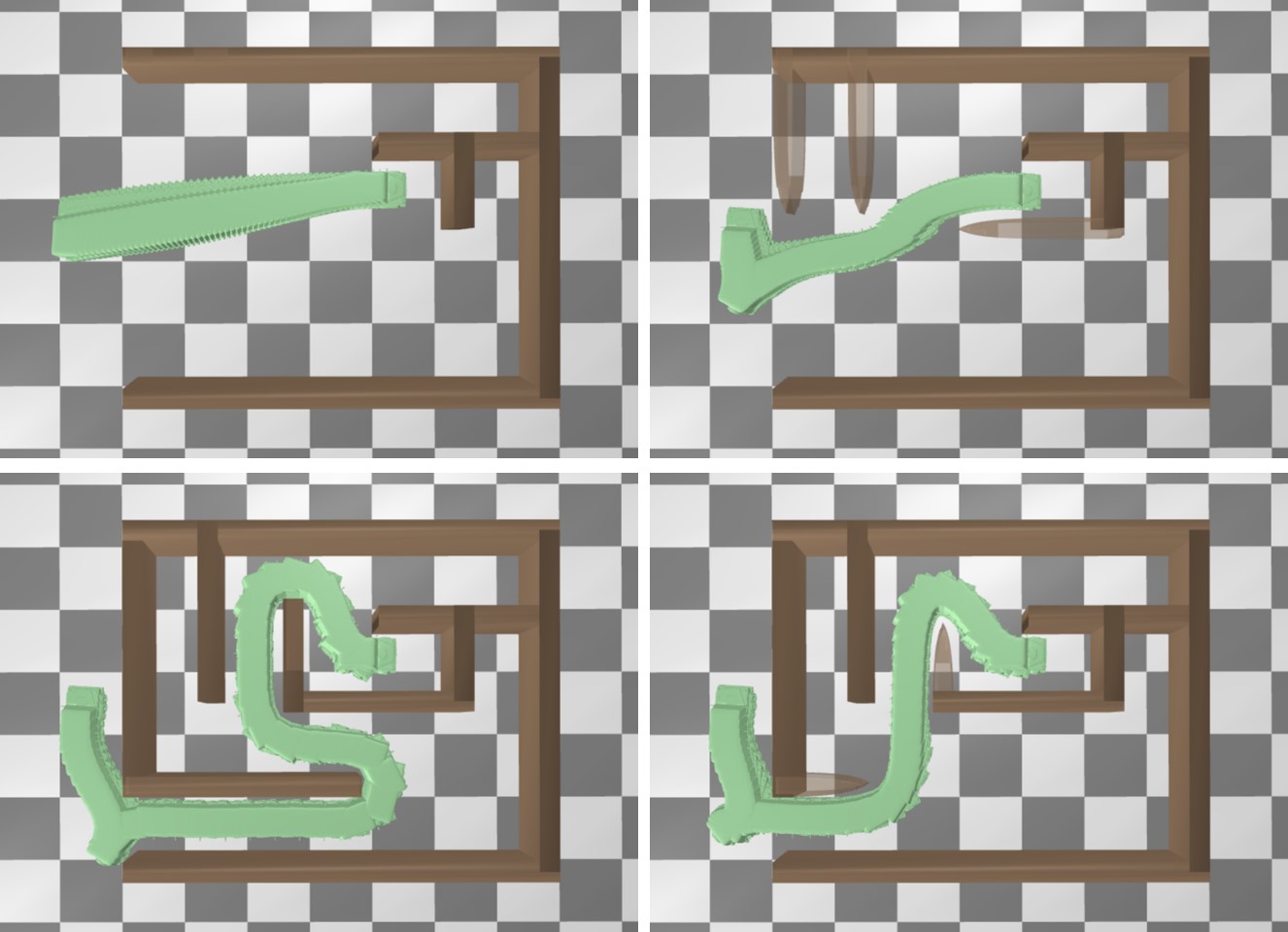}
    \caption{Process of the path planning optimization of a nonholonomic system 
 in a maze (clock-wise).}
    \label{fig:nonholonomic}
\end{figure}

Nonholonomic navigation of maze has been studied using various approaches such as RRT-based nonholonomic planning \cite{palmieri2016rrt}. However, once the volume of the vehicle is considered, the corridors in the configuration space become extremely narrow, causing a significant loss of scalability of sampling-based approaches.
Our methodology can augment nonholonomic constraints within a general optimization framework and address the narrow corridor problem through the collision constraint interpolation. 
The result of the optimization is illustrated in Fig. \ref{fig:nonholonomic}. Starting from an environment with enlarged free space, the path of the vehicle is effectively optimize to navigate through the maze.

\section{Conclusions} \label{sec:conclusion}

We present a path planning framework specifically designed to address the challenges posed by narrow passage, but it has several limitations.
Firstly, objects can only be added when contained in a leaf set, which may limit the applicability of the framework in environments with complex topologies.
Moreover, the collision constraint interpolation incurs additional overhead due to the convex decomposition of the environment. Additionally, as the problem is solved iteratively by increasing the interpolation variable $\alpha$, this method can be time-consuming in scenarios where extremely narrow passages are not present.
 Finally, the optimization-based nature of our methodology risks getting stuck in local minima. Future work includes analyzing homotopy and extending the framework to more general and complex environments.

\bibliographystyle{unsrt}
\bibliography{reference}
\clearpage

\end{document}